\documentclass[letterpaper, 10 pt, conference]{ieeeconf}  % Comment this line out if you need a4paper

\IEEEoverridecommandlockouts                              % This command is only needed if 
\usepackage{fancyhdr}
\fancypagestyle{firstpage}{%
  \lhead{\footnotesize 2022 IEEE/RSJ International Conference on Intelligent Robots and Systems (IROS 2022). Preprint. Accepted June 2022. \\
  2022 ©IEEE. Personal use of this material is permitted. Permission from IEEE must be obtained for all other uses, in any current or future media, including reprinting/republishing this material for advertising or promotional purposes, creating new collective works, for resale or redistribution to servers or lists, or reuse of any copyrighted component of this work in other works.}
  \rhead{}
}

\usepackage{graphics} % for pdf, bitmapped graphics files
\usepackage{amssymb}  % assumes amsmath package installed
\usepackage{textglos}
\usepackage[utf8]{inputenc} % allow utf-8 input
\usepackage[T1]{fontenc}    % use 8-bit T1 fonts
\usepackage{hyperref}       % hyperlinks
\usepackage{url}            % simple URL typesetting
\PassOptionsToPackage{hyphens}{url}\usepackage{hyperref}
\usepackage[hyphenbreaks]{breakurl}
\usepackage{booktabs}       % professional-quality tables
\usepackage{amsfonts}       % blackboard math symbols
\usepackage{nicefrac}       % compact symbols for 1/2, etc.
\usepackage{microtype}      % microtypography
\usepackage{lipsum}
\usepackage{tabularx} % extra features for tabular environment
\usepackage{amsmath}  % improve math presentation
\usepackage{graphicx,dblfloatfix} % takes care of graphic including machinery
\usepackage[letterpaper, left=0.625in, right=0.625in, bottom=0.8in, top=0.7in]{geometry} % decreases margins
\usepackage{cite} % takes care of citations
\usepackage{titlesec}
\usepackage{mathtools, amssymb, nccmath}
\usepackage{wasysym}
\usepackage{algorithm}
\usepackage{algorithmic}
\usepackage{multicol}
\let\labelindent\undefined
\usepackage{enumitem}
\usepackage{float}
\usepackage{mathtools}
\usepackage{amsmath,amsfonts,amssymb}
\usepackage{cmll}
\usepackage{tensor}
\usepackage{booktabs}
\usepackage{tabularx}
\usepackage{makecell}
\usepackage{tablefootnote}
\usepackage{threeparttable}
\usepackage{multirow}

\allowdisplaybreaks[1]

\usepackage{setspace}
\usepackage[font=small,skip=5pt]{caption}

\usepackage[dvipsnames]{xcolor}

\newcolumntype{Z}{ >{\centering\arraybackslash}X }

\newcommand{\crff}{\,\overline{\!\times\!}{}^{\,*}}

\newcommand{\mysp}{.7ex}

\newcommand{\M}{\mathcal{M}}
\newcommand{\F}{\boldsymbol{F}}
\newcommand{\I}{\mathcal{I}}

\newcommand{\T}{^\top}
\newcommand{\Tten}{^{\widetilde{\top}}} % transpose 1-2 for tensor
\newcommand{\Rten}{^{\widetilde{\mathrm{R}}}} % transpose 2-3 for tensor
\newcommand{\RTten}{^{\widetilde{\mathrm{R}},\widetilde{\!\top\!}}} % transpose 2-3 for tensor

\newcommand{\R}{\mathbb{R}}
\newcommand{\C}{\vC}

\newcommand{\rmvec}[1]{\boldsymbol{#1}}
\newcommand{\greekvec}[1]{\boldsymbol{#1}}
\newcommand{\B}{\boldsymbol{B}{}}
\renewcommand{\C}{\boldsymbol{C}}
\renewcommand{\M}{\boldsymbol{M}}
\newcommand{\A}{\boldsymbol{A}{}}

\renewcommand{\I}{\boldsymbol{I}{}}
\newcommand{\IC}[1]{\I_{#1}^{C}}
\newcommand{\BC}[1]{\B_{#1}^{C}}
\newcommand{\BCT}[1]{\B_{#1}^{C,\top}}

\newcommand{\g}{\rmvec{g}}
\newcommand{\q}{\rmvec{q}}

\renewcommand{\u}{\rmvec{u}}

\newcommand{\U}{\rmvec{U}}
\newcommand{\V}{\rmvec{V}}
\usepackage{framed}

\newcommand{\f}{\rmvec{f}}

\renewcommand{\v}{\rmvec{v}}

\renewcommand{\a}{\rmvec{a}}

\newcommand{\qd}{\dot{\q}}
\newcommand{\qdd}{\ddot{\q}}

\newcommand{\Psibar}{\greekvec{\Psi}}
\newcommand{\Psibardot}{\dot{\Psibar}}
\newcommand{\Psibarddot}{\ddot{\Psibar}}

\newcommand{\psibar}{\greekvec{\psi}}
\newcommand{\psibardot}{\,\dot{\!\psibar}}
\newcommand{\psibarddot}{\,\ddot{\!\psibar}}

\newcommand{\taubar}{\greekvec{\tau}}

\newcommand{\red}[1]{{\color{red}#1}}

\newcommand{\blue}[1]{{\color{blue}#1}}

\newcommand{\phibar}{\boldsymbol{s}}
\newcommand{\Phibar}{\boldsymbol{S}}
\newcommand{\phibardot}{\,\dot{\!\phibar}}
\newcommand{\Phibardot}{\,\dot{\!\Phibar}}

\usepackage{scalerel}
\newcommand{\timesM}{{\tilde{\smash[t]{\times}}}}
\newcommand{\timesfM}{\timesM {}^*}
\newcommand{\crffM}{\scaleobj{.87}{\,\tilde{\bar{\smash[t]{\!\times\!}}{}}}{}^{\,*}}

\newcommand{\timesf}{{\times}^*}

\newcommand{\dtaudqSO}[3]{\frac{\partial ^{2} \taubar_{#1}}{\partial \q_{#2} \partial \q_{#3}}}

\newcommand{\dtaudqdSO}[3]{\frac{\partial ^{2} \taubar_{#1}}{\partial \qd_{#2} \partial \qd_{#3}}}

\newcommand{\dtaudqdMSO}[3]{\frac{\partial ^{2} \taubar_{#1}}{\partial \q_{#2} \partial \qd_{#3}}}

\newcommand{\Bicphi}[2]{\mathcal{B}_{#1}^C[\Phibar_{#2}]}
\newcommand{\Bicpsiidot}[2]{\mathcal{B}_{#1}^{C}\big[\Psibardot_{#2}\big]}

\newcommand{\Bicphip}[1]{\B_{#1}^C[\phibar_{#1,p}]}
\newcommand{\dMdq}[3]{\frac{\partial  \M_{#1,#2}}{\partial \q_{#3} }}

\usepackage{tikz}
\usetikzlibrary{matrix,calc}

\setlist[enumerate]{wide=0pt, widest=99,leftmargin=25pt, labelsep=*}

\newcommand{\Ff}{\F}
\newcommand{\Uu}{\U}
\newcommand{\Uv}{\V}

\newcommand{\appref}[1]{App.~\ref{#1}}

\title{\Large \bf Analytical Second-Order Partial Derivatives of Rigid-Body Inverse Dynamics\\[-2ex]
}

\author{ Shubham Singh$^{1}$, Ryan P. Russell$^{1}$ and Patrick M.~Wensing$^{2}$% \prec-this % stops a space
\thanks{$^{1}$ Aerospace Engineering, The University of Texas at Austin, TX-78751, USA. \href{mailto:singh281@utexas.edu}{singh281@utexas.edu}, \href{mailto:ryan.russell@utexas.edu}{ryan.russell@utexas.edu}}
       \thanks{$^{2}$ Aerospace \& Mechanical Engineering, University of Notre Dame, IN-46556, USA. \href{mailto:pwensing@nd.edu}{pwensing@nd.edu}
       } %
       \thanks{This work was supported in part by the National Science Foundation grants CMMI-1835013 and CMMI-1835186.}
       }

\begin{document}

\maketitle
\thispagestyle{firstpage}
\pagestyle{plain}

%%%%%%%%%%%%%%%%%%%%%%%%%%%%%%%%%%%%%%%%%%%%%%%%%%%%%%%%%%%%%%%%%%%%%%%%%%%%%%%%

\begin{abstract}
Optimization-based robot control strategies often rely on first-order dynamics approximation methods, as in iLQR. Using second-order approximations of the dynamics is expensive due to the costly second-order partial derivatives of the dynamics with respect to the state and control. Current approaches for calculating these derivatives typically use automatic differentiation (AD) and chain-rule accumulation or finite-difference.
In this paper, for the first time, we present analytical expressions for the second-order partial derivatives of inverse dynamics for open-chain rigid-body systems with floating base and multi-DoF joints. A new extension of spatial vector algebra is proposed that enables the analysis. A recursive algorithm with complexity of $\mathcal{O}(Nd^2)$ is also provided where $N$ is the number of bodies and $d$ is the depth of the kinematic tree. A comparison with AD  in  CasADi shows speedups of 1.5-3$\times$ for serial kinematic trees with $N> 5$, and a C++ implementation shows runtimes of $\approx$51$\mu s$ for a quadruped.
\end{abstract}

\section{Introduction}
In recent years, optimization-based methods have become popular for robot motion generation and control. Although full second-order (SO) optimization methods offer superior convergence properties, most work has focused on using only the first-order (FO) dynamics approximation, such as in iLQR~\cite{tassa2012synthesis, koenemann2015whole}.
Differential Dynamic Programming (DDP)~\cite{mayne1966second} is a use-case for full SO optimization that has gained wide interest for robotics applications~\cite{tassa2014control,chatzinikolaidis2021trajectory,mastalli2020crocoddyl,li2020hybrid}. Variants of DDP using multiple shooting~\cite{pellegrini2020multiple} and parallelization~\cite{plancher2018performance} have been developed for computational and numerical improvements. %improve convergence and robustness while reducing the computational time. 

The state-of-the-art for including SO dynamics derivatives in trajectory optimization is the work by Lee et al.~\cite{lee2005newton}, where they use forward chain-rule expressions (i.e., recursive derivative expressions) to calculate the SO partial derivatives of joint torques with respect to joint configuration and joint rates for revolute and prismatic joint models. 
%\red{Chain-rule expressions are inherently recursive in nature, since they relate derivatives at one step in the algorithm to those at the previous.}
Nganga and Wensing~\cite{nganga2021accelerating} presented a method for getting the SO directional derivatives of inverse dynamics as needed in the backward pass of DDP by employing reverse mode Automatic Differentiation (AD) and a modified version of the Recursive Newton Euler Algorithm (RNEA)~\cite{Featherstone08}. Although this strategy avoids the need for the full SO partial derivatives of RNEA, it lacks the opportunity for parallel computation across the trajectory. Full SO partial derivatives calculation, on the other hand, can be easily parallelized to accelerate the backward pass of DDP. Other trajectory optimization schemes aside from DDP may also benefit from the full SO partials.

AD tools depend on forward or reverse chain-rule accumulation and can suffer from high memory requirements~\cite{kowarz2006optimal}. %and computational time \red{citation?}. 
Finite-difference methods, on the other hand, can be parallelized, but suffer from low accuracy. Such inaccurate Jacobian and Hessian approximations during optimization often lead to ill-conditioning, and poor convergence~\cite{lee2005newton}.

% Mueller~\cite{muller} presents recursive algorithms for SO time derivatives of inverse dynamics, often required for robots with elastic actuators.
 
 Although significant work has been done for computing FO partial derivatives of inverse/forward dynamics~\cite{singh2022efficient,jain1993linearization,ayusawa2018comprehensive,carpentier2018analytical}, the literature still lacks analytical SO partial derivatives of rigid-body dynamics. This is mainly due to the tensor nature of SO derivatives, and the lack of established tools for working with dynamics tensors. 
 The main contribution of this paper is to extend the FO derivatives of inverse dynamics (ID) presented in Ref.~\cite{singh2022efficient} to SO derivatives w.r.t the joint configuration ($\q$),  velocity vector ($\qd$), and joint acceleration ($\qdd$) for multi-Degree-of-Freedom (DoF) joints modeled with Lie groups. Our method departs from the chain-rule approach used in~\cite{lee2005newton}, and thus gains additional efficiency in the associated algorithm.  For this purpose, we contribute an extension of Featherstone's spatial vector algebra tools~\cite{Featherstone08} for tensor use. The SO derivatives algorithm herein could also be parallelized (e.g., for use with GPUs), building on the FO case in Ref.~\cite{plancher2021accelerating}. 
 
 In the following sections, Spatial Vector Algebra (SVA) is reviewed for dynamics analysis, followed by an extension of SVA for use with second-order derivative tensors. Then, the algorithm is developed for the SO partial derivatives of ID. The resulting expressions and algorithm are very complicated, but are provided in the form of open-source algorithm, with full derivation in Ref.~\cite{arxivss_SO} while keeping the current paper self-contained. The  performance of this new SO algorithm is compared to AD using CasADi~\cite{andersson2019casadi} in {\sc Matlab}.
 
\section{Rigid-Body Dynamics Background}
\label{sec:pd_rbd}

\noindent {\bf Rigid-Body Dynamics}: 
 For a rigid-body system with $n$-dimensional configuration manifold $\mathcal{Q}$, the state variables are the configuration $\q \in \mathcal{Q}$ and the generalized velocity vector $\qd\in\mathbb{R}^n$, while the control variable is the generalized force/torque vector $\taubar \in \mathbb{R}^n$. With this convention, $\qd$ uniquely specifies the time rate of change for $\q$ without strictly being its time derivative. The Inverse Dynamics (ID), is given by 
 \begin{align}
    \taubar &= \M(\q)\qdd+ \C(\q,\qd)\qd + \g(\q)
    \label{inv_dyn} \\
&= \textrm{ID}({\rm model},\q,  \qd, \qdd)
    \label{inv_dyn_model}
\end{align}
where $\M \in \R^{n \times n}$ is the mass matrix, $\C \in \R^{n \times n}$ is the Coriolis matrix, and  $\g \in \R^{n}$ is the vector of generalized gravitational forces. %, and $n$ is the DoF of the system. 
%For a fixed $\q$, and $\qd$, ID calculates $\taubar$ for a given $\qdd$ (Eq.~\ref{inv_dyn}). 
An efficient $\mathcal{O}(N)$ algorithm for ID is the RNEA \cite{orin1979kinematic,Featherstone08}. 
%Analytical expressions for partial derivatives of ID w.r.t $\q$, $\qd$ were presented in Ref.~\cite{singh2022efficient}, among other similar formulations~\cite{jain1993linearization,ayusawa2018comprehensive}. 
 
%\section{Spatial Vector Algebra (SVA)}
%\label{iden_defn}

\noindent {\bf Notation:} Spatial vectors are 6D vectors that combine the linear and angular aspects of a rigid-body motion or net force~\cite{Featherstone08}. Cartesian vectors are denoted with lower-case letters with a bar ($\bar{v}$), spatial vectors with lower-case bold letters (e.g., $\a$), matrices with capitalized bold letters (e.g., $\boldsymbol{A}$), and tensors with capitalized calligraphic letters (e.g., $\mathcal{A}$). Motion vectors, such as velocity and acceleration, belong to a 6D vector space denoted $M^{6}$. Force-like vectors, such as force and momentum, belong to another 6D vector space $F^{6}$. Spatial vectors are usually expressed in either the ground coordinate frame or a body coordinate (local) frame. For example, the spatial velocity ${}^{k}\v_{k} \in M^{6}$ of a body $k$ expressed in the body frame is
%\begin{align}
${}^{k}\v_{k} = \begin{bmatrix}
           {}^{k}\bar{\omega}_{k}\T &
           {}^{k}\bar{v}_{k}\T
         \end{bmatrix}\T$
%\end{align}
where ${}^{k}\bar{\omega}_{k}$ $\in \mathbb{R}^{3}$ is the angular velocity expressed in a coordinate frame fixed to the body, while ${}^{k}\bar{v}_{k}$ $\in \mathbb{R}^{3}$ is the linear velocity of the origin of the body frame.
When the frame used to express a spatial vector is omitted, the ground frame is assumed.

 A spatial cross product between motion vectors ($\v$,$\u$), written as $(\v \times) \u$, is given by Eq.~\ref{cross_defn}. This operation gives the time rate of change of $\u$, when $\u$ is moving with a spatial velocity $\v$.  
For a Cartesian vector $\bar{\omega}$, $\bar{\omega} \times$ is the 3D cross-product matrix. A spatial cross product between a motion and a force vector is written as $(\v \times^*) \f$, as defined by Eq.~\ref{cross_f_defn}.
\begin{multicols}{2}
  \begin{align}
    \v \times = \begin{bmatrix}
       \bar{\omega}  \times & \bf{0}  \\
       \bar{v} \times &  \bar{\omega} \times
    \end{bmatrix}
    \label{cross_defn}
\end{align}
  \begin{align}
    \v \times^* = \begin{bmatrix}
        \bar{\omega} \times & \bar{v} \times  \\
        \bf{0} &  \bar{\omega} \times
    \end{bmatrix}
    \label{cross_f_defn}
\end{align}
\end{multicols}
\noindent An operator $\crff$ is  defined by swapping the order of the cross product, such that $(\f \crff) \v = (\v \times^*) \f$~\cite{echeandia2021numerical}.  Further introduction to  SVA is provided in Ref.~\cite{Featherstone08}.

\begin{figure}[tb]
 \centering
 \includegraphics[width=.80 \columnwidth]{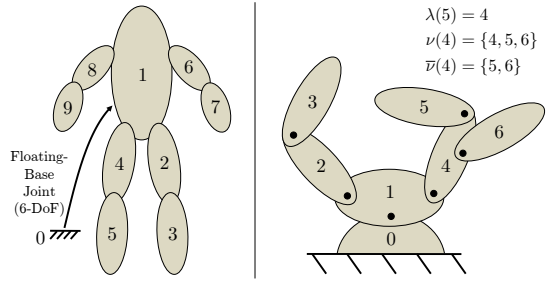}
 \caption{Convention examples with floating- and fixed-base systems.}
 \label{link_fig}
 \vspace{-5px}
 \end{figure}

\noindent {\bf Connectivity:} An open-chain kinematic tree (Fig.~\ref{link_fig}) is considered with $N$ links connected by joints, each with up to 6 DoF. Body $i$'s parent toward the root of the tree is denoted as $\lambda(i)$, 
%while the set $\nu(i)$ denotes the set of bodies in the subtree rooted at body $i$. %, while $\subtreeb(i) = \subtree(i) \setminus \{i\}$. %denotes the set of bodies in $\subtree(i)$ excluding the body $i$ . 
and we define $i \preceq j$ if body $i$ is in the path from body $j$ to the root.  Joint $i$ is defined as the connection between body $i$ and its predecessor.

We consider joints whose configurations form a sub-group of the Lie group \textit{SE}(3). For a prismatic joint, the configuration and rate are represented by $\q_{i},\qd_{i}\in \mathbb{R}$, while for a revolute joint, $\q_{i} \in SO(2)$, and $\qd_{i} \in \mathbb{R}$ gives the rotational rate of the joint. For a spherical joint, $\q_{i} \in SO(3)$, and $\qd_{i} = {}^i \bar{\omega}_{i/\lambda(i)} \in \R^3$ gives relative angular velocity between neighboring bodies. 
For a 6-DoF free motion joint, $\q_{i} \in SE(3)$, and $\qd_{i} = {}^i \v_{i/\lambda(i)} \in \R^6$. %Thus, $\qd$ is not strictly the time derivative of $\q$ in general.

The spatial velocities of the neighbouring bodies in the tree are then related by the recursive expression $\v_{i} = \v_{\lambda(i)} + \Phibar_{i} \qd_{i}$, where $\Phibar_{i} \in \mathbb{R}^{6\times n_i}$ is the joint motion subspace matrix for joint $i$ \cite{Featherstone08} with $n_i$ its number of DoFs. % and $\qd_{i}$ the joint rates for joint $i$.  %For a single DoF revolute joint, $\Phibar_{i} \in \mathbb{R}^{6}$, while for a 6-DoF free-motion joint, $\Phibar_{i} \in \mathbb{R}^{6 \times 6}$. 
 The velocity $\v_{i}$ can also be written in an analytical %(i.e., non-recursive) 
 form as the sum of joint velocities over predecessors as $\v_{i} = \sum_{l \preceq i} \Phibar_{l} \qd_{l}$.
 The derivative of the joint motion subspace matrix in local coordinates (often denoted $\,\mathring{\!\Phibar}_{i}$~\cite{Featherstone08}) is assumed to be zero.
The quantity $\Phibardot{}_{i} = \v_{i} \times \Phibar_{i}$ signifies the rate of change of $\Phibar_{i}$ due to the local coordinate system moving.

\noindent {\bf Dynamics:} The spatial equation of motion~\cite{Featherstone08} is given for body $k$ as 
$\f_{k}= \I_{k}\a_{k} + \v_{k} \times^*\I_{k} \v_{k}$, 
where $\f_{k}$ is the net spatial force on body $k$, $\I_{k}$ is its spatial inertia~\cite{Featherstone08}, and $\a_{k}$ is its spatial acceleration. %Instead of treating gravity as an external force, a common trick is to accelerate the base upwards opposite of the gravitational acceleration ($\a_{0} = -\a_{g}$).
 The development in Ref.~\cite{singh2022efficient} presents two additional spatial motion  quantities $\Psibardot$, $\Psibarddot$ (Eq.~\ref{psidot_psidotdot_defn}) essential to the derivation in this paper. The quantities $\Psibardot_{j}$ and $\Psibarddot_{j}$ represent the time-derivative of $\Phibar_{j}$, and $\Psibardot_{j}$, respectively, due to joint $j$'s predecessor $\lambda(j)$ moving.
\begin{align}
\begin{split}
  \Psibardot_{j} = & \v_{\lambda(j)} \times \Phibar_j \\
 \Psibarddot_{j} = & \a_{\lambda(j)} \times \Phibar_{j} + \v_{\lambda(j)} \times \Psibardot_{j}
\end{split}
\label{psidot_psidotdot_defn}
\end{align}

The analytical expressions for the FO partial derivatives of ID w.r.t $\q$ and $\qd$~\cite{singh2022efficient} are given below,  with other similar formulations in~\cite{jain1993linearization,ayusawa2018comprehensive}. Here $\taubar_{i}$ represents the joint torques/forces for joint $i$ from Eq.~\ref{inv_dyn}. The quantity $\f{}_{i}^{C} = \sum_{k \succeq i} \f_{k}$ is the composite spatial force transmitted across joint $i$, and $\IC{i}$ is the composite rigid-body inertia of the sub-tree rooted at body $i$, given as $\I{}_{i}^{C} = \sum_{k \succeq i} \I_{k}$. The quantity $\B_{k}$ is a body-level Coriolis matrix \cite{singh2022efficient,echeandia2021numerical},
\begin{equation}
\B_{k} = \scaleobj{.75}{ \frac{1}{2}} [(\v_k \times^* )\I_{k} - \I_{k} (\v_k \times) + (\I_{k} \v_k ) \crff]
\label{bk_term_defn}
\end{equation}
while $\BC{i}$ is its composite given by $\B{}_i^C = \sum_{k\succeq i} \B_k$. The next sections extend Eqs.  \ref{dtau_dq}-\ref{dtau_dqd} for SO derivatives of ID.
% If instead of $\v_{k}$, the body-Coriolis term $\B_k$ is evaluated using some other spatial quantity $\m$ as an argument, then it is denoted with a square bracket as $\B_{k}[\m]$, and defined as:
% \begin{equation}
%     \B_{k}[m] = \frac{1}{2} \big(\big(\m \times^* \big)\I_{k} - \I_{k} \big(\m \times   \big) + \big(\I_{k} \m \big) \crff    \big)
%     \label{bkm_term_defn}
% \end{equation}
% Composite term for $\B_{k}[m]$ is $\BC{k}[m] = \sum_{l \succeq k} \B_{l}[m]$. 
%

\begin{subequations}
 \label{dtau_dq}
\begin{eqnarray}
     \frac{\partial {\taubar_{i}}}{\partial \q_{j}} &=&  { \Phibar_{i}\T \big[ 2  \B_{i}^{C} \big] \Psibardot{}_{j} + \Phibar_{i}\T  \I_{i}^{C} \Psibarddot{}_{j} }, (j \preceq i)
          \label{dtau_dq_1} 
\\
\frac{\partial {\taubar_{j}}}{\partial {\q_{i}}} &=&  {\Phibar_{j}\T [ 2 \B_{i}^{C} \Psibardot_{i} +  \I_{i}^{C}  \Psibarddot_{i}+(\f_{i}^{C} )\crff \Phibar_{i}  ], (j \prec i) }~~ 
     \label{dtau_dq_2}
\end{eqnarray}
\end{subequations}
\vspace{-10px}
\begin{subequations}
\label{dtau_dqd}
\begin{eqnarray}
         \frac{\partial {\taubar_{i}}}{\partial {\dot{\q}_{j}}} &=& {\Phibar_{i}\T \Big[2 \B_{i}^{C} \Phibar_{j} +  \I_{i}^{C} (    \Psibardot_{j} + \Phibardot_{j} ) \Big],  (j \preceq i) }
         \label{dtau_dqd_1}
\\
    \frac{\partial {\taubar_{j}}}{\partial {\dot{\q}_{i}}}  &=& {\Phibar_{j}\T \Big[2 \B_{i}^{C} \Phibar_{i} +  \I_{i}^{C} (    \Psibardot_{i} + \Phibardot_{i} ) \Big], (j \prec i)}~~~~~~~~\,
         \label{dtau_dqd_2}
     \end{eqnarray}
    \end{subequations}

%In the following sections, we extend the partial derivatives of ID to SO using Eq.~\ref{dtau_dq_1}-\ref{dtau_dqd_2}.

\section{Extending SVA For Tensorial Use}
\label{sva_tensor}
\begin{figure}
\center
\includegraphics[width=6cm]{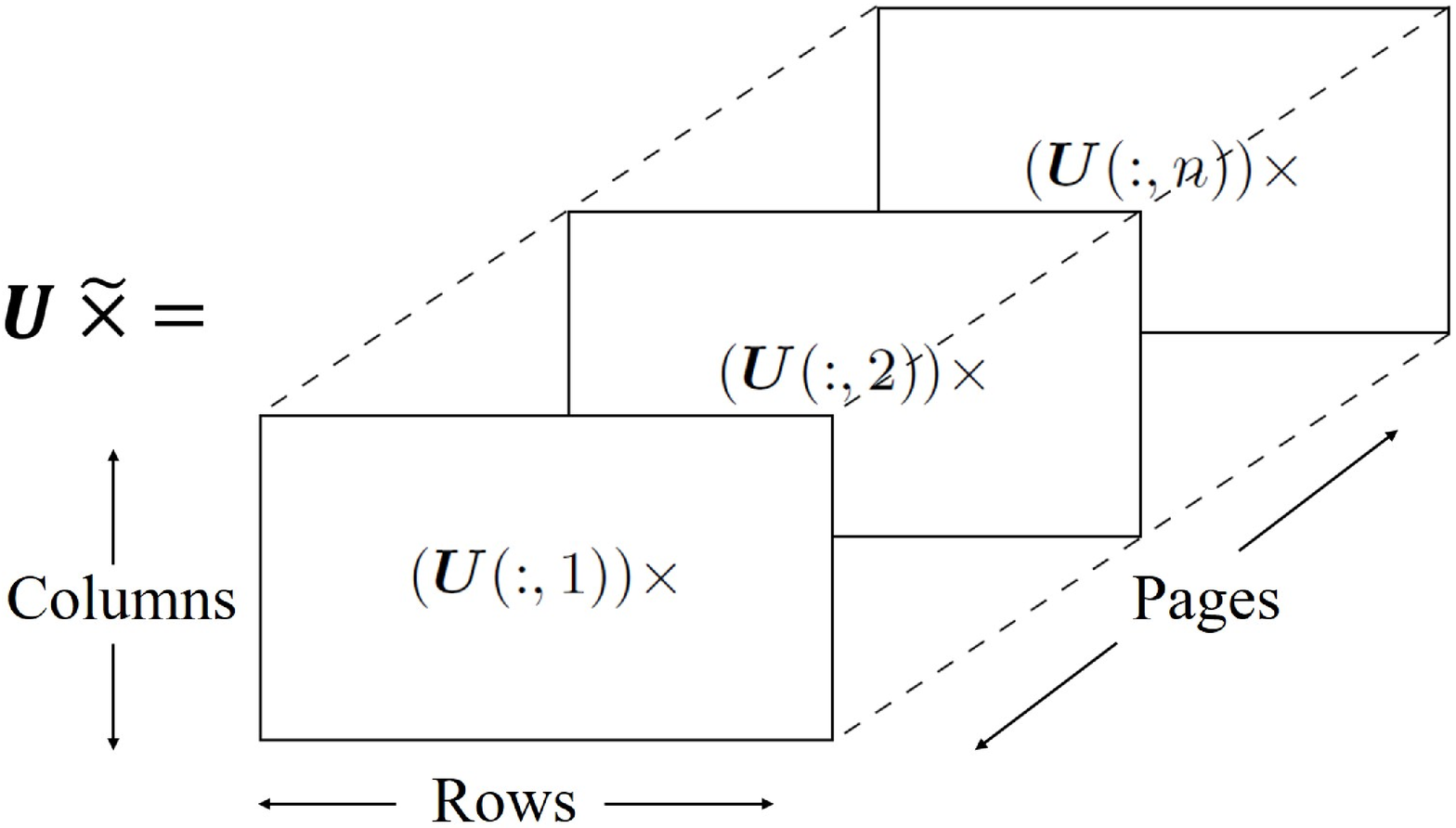}
\caption{$\timesM$ operates on each column of a $\U \in M^{6\times n}$ spatial matrix to create a third-order tensor. Each rectangular box is a 2D matrix} \label{fig:crossM}
\vspace{-5px}
\end{figure}

The motion space $M^{6}$~\cite{Featherstone08} is extended to a space of spatial-motion matrices $M^{6\times n}$, where each column of such a matrix is a usual spatial motion vector.
For any $\U \in M^{6 \times n}$ a new spatial cross-product operator $ \timesM$ is considered and defined by applying the usual spatial cross-product operator ($\times$) to each column of $\U$. The result is a third-order tensor (Fig.~\ref{fig:crossM}) in $\R^{6 \times 6 \times n}$ where each $6 \times 6$ matrix in the 1-2 dimension is the original spatial cross-product operator on a column of $\U$. %Therefore, the operator $\timesM$ maps a spatial-motion matrix $M^{6 \times n_{2}}$ to a third-order tensor as:

% \begin{equation}
%   \timesM : M^{6\times n_{1}} \cdot M^{6 \times n_{2}} \xrightarrow[]{} M^{6 \times n_{2} \times n_{1}} \\
% \end{equation}
% where, $\cdot$ represents a tensor-matrix product defined in Sec.~\ref{ten_mat_product_sec} between the $6 \times 6 \times n_{1}$ tensor generated by using $\timesM$ operator on the $ M^{6\times n_{1}}$ spatial-motion matrix, and the $M^{6 \times n_{2}}$ spatial-motion matrix.

Given two spatial motion matrices, $\U \in M^{6 \times n_u}$ and $\V \in M^{6 \times n_v}$ , we can now define a cross-product operation between them as $(\U \timesM) \V \in M^{6 \times n_v \times n_u}$ via a tensor-matrix product. Such an operation, denoted as $\mathcal{Z} = \mathcal{A} \B$ is defined as:
\begin{equation}
    \mathcal{Z}_{i,j,k} = \sum_{\ell}\mathcal{A}_{i,\ell,k}\\B_{\ell,j}
\end{equation}
for any tensor $\mathcal{A}$ and suitably sized matrix $\B$. Thus, the $k$-th page, $j$-th column of $\U \timesM \V$ gives the cross product of the $k$-th column of $\U$ with the $j$-th column of $\V$.

In a similar manner, consider a spatial force matrix $\F\in F^{6 \times n_f}$. Defining $(\V\timesfM)$ in an analogous manner to in Fig.~\ref{fig:crossM} allows taking a cross-product-like operation $\V\timesfM \F$. Again, analogously, we consider a third operator $(\F\crffM)$ that provides  $\V\timesfM \F = \F\crffM \V$. In each case, the tilde indicates the spatial-matrix extension of the usual spatial-vector cross products.

For later use, the product of a matrix $\B \in \R^{n_{1} \times n_{2}}$, and a tensor $\mathcal{A}\in \R^{n_{2} \times n_{3} \times n_{4}}$, likewise results in another tensor, denoted as $\mathcal{Y} = \B\mathcal{A}$, and defined as:

\begin{equation}
    \mathcal{Y}_{i,j,k} = \sum_{\ell}\B_{i,\ell}\mathcal{A}_{\ell,j,k}
\end{equation}

Two types of tensor rotations are defined for this paper:
\begin{enumerate}
    \item  $\mathcal{A}\Tten$: Transpose along the 1-2 dimension. This operation can also be understood as the usual matrix transpose of each matrix (e.g., in Fig.~\ref{fig:crossM}) moving along pages of the tensor. If $ \mathcal{A}\Tten = \mathcal{B}$, then $\mathcal{A}_{i,j,k} = \mathcal{B}_{j,i,k}$.\\[-1.7ex]
    \item  $\mathcal{A}\Rten$: Rotation of elements along the 2-3 dimension. If $ \mathcal{A}\Rten = \mathcal{B}$, then $\mathcal{A}_{i,j,k} = \mathcal{B}_{i,k,j}$.
\end{enumerate}
Another rotation ($\RTten$) is a combination of  $(\Rten)$ followed by $(\Tten)$. For example, if $\mathcal{A}\RTten = \mathcal{B}$, then $\mathcal{A}_{i,j,k} = \mathcal{B}_{k,i,j}$.

Properties of the operators are given in Table~\ref{tab:SVM_identites}. These properties naturally extend spatial vector properties~\cite{Featherstone08}, but with the added book-keeping required from using tensors. For example, the spatial force/vector cross-product operator $\timesf$ satisfies $\v\timesf = -\v\times \T$. Property M1 provides the matrix analogy for the spatial matrix operator $\timesfM$. %Property M8 is the spatial matrix analogy of the spatial vector property $\u \times \v = -\v \times \u$, where $\u$ and $\v$ are spatial vectors.

\begin{table}[t]
\center 
\fbox{
\begin{minipage}{.88\columnwidth}
{\small
\begin{enumerate}[label=M{{\arabic*}})]
 \setlength\itemsep{.15em}
    \item \label{m1} $\U \timesfM = -(\U \timesM)\Tten$ 
    \item \label{m2} $-\Uv\T (\Uu \timesfM) = (\Uu \timesM \Uv)\Tten $
    \item \label{m3} $-\Uv\T (\Uu \timesfM) \Ff= (\Uu \timesM \Uv)\Tten \Ff$
    \item \label{m4} $ (\U \timesM \v) = - \v \times \U $ 
    \item \label{m5} $\U \timesfM \Ff = ( \Ff \crffM \U)\Rten$ 
    \item \label{m6} $\Ff \crffM \U = (\U \timesfM \Ff)\Rten $ 
    \item \label{m7} $(\lambda \U) \timesM = \lambda(\U \timesM)$
    \item \label{m8} $\Uu \timesM \Uv = - (\Uv \timesM \Uu)\Rten $ 
    \item \label{m9} $(\v \times \U) \timesM = \v \times \U \timesM - \U \timesM \v \times$ 
    \item \label{m10} $(\v \times \U) \timesfM = \v \times^* \U \timesfM - \U \timesfM \v \times^*$
    \item \label{m11} $(\U \timesfM \v) \crffM = \U \timesfM \v \crff - \v \crff \U \timesM$
    \item \label{m12} $(\U \timesfM \Ff)\Tten = -\Ff\T (\U \timesM)$
    \item \label{m13} $\Uv\T (\Uu \timesfM \Ff) = (\Uv \timesM \Uu)\RTten \Ff=  (\Ff\T (\Uv \timesM \Uu)\Rten)\Tten $
    \item \label{m14} $ \v \times^* \Ff = \Ff \crffM \v$
    \item \label{m15} $ \f \crff \U = \U \timesfM \f$
     \item \label{m16} $\Uv\T (\Uu \timesfM \Ff)\Rten = \big[(\Uv \timesM \Uu)\RTten \Ff\big]\Rten$
     \item \label{m17} $\Uv\T (\Uu \timesfM \Ff)\Rten = -[\Uu\T (\Uv \timesfM \Ff)\Rten]\Tten $ 
      \item \label{m18} $\Uv \T (\Uu \timesfM \Ff)\Rten = [ \Uv\T (\Uu \timesfM \Ff)]\Rten$ 
     \item \label{m19} $(\B\mathcal{Y})\Tten = \mathcal{Y}\Tten \B\T$
      %\item \label{m20} $(\mathcal{Y}+\mathcal{Z})\Tten = \mathcal{Y}\Tten+\mathcal{Z}\Tten$
       %\item \label{m21} $(\mathcal{Y}+\mathcal{Z})\Rten = \mathcal{Y}\Rten+\mathcal{Z}\Rten $
\end{enumerate}
}
\end{minipage}
}
\caption{Spatial Matrix Algebra Identities: $\v \in M^{6}$,$\f \in F^{6}$, $\U \in M^{6 \times n}$, $\Ff \in F^{6 \times m} $, $\Uv \in M^{6 \times l}$, $\B \in \R^{n_{1} \times n_{2}}$, $\mathcal{Y} \in \R^{n_{2} \times n_{3} \times n_{4}}$.}
\label{tab:SVM_identites}
\vspace{-15px}
\end{table}

\section{Second-Order Derivatives of ID}
\label{so_partials_sec}
\subsection{Preliminaries}
\label{prem_sec}
The SO partial derivative of joint torque/force is also referred to as a dynamics Hessian tensor. Blocks of this rank 3 tensor are written in a form $\frac{\partial^2 \taubar_{i}}{\partial \u_{j} \partial \u_{k}}$, which signifies taking partial derivative of $\taubar_{i}$ w.r.t $\u_{j}$, followed by $\u_{k}$. The variables $\u_{j}$ and $\u_{k}$ can either be the joint configuration ($\q_j$, $\q_k$), joint velocity ($\qd_j$, $\qd_k$), or joint acceleration ($\qdd_j$, $\qdd_k$). Many of these second-order partials are zero, limiting the cases to be considered. From Eq.~\ref{inv_dyn}, the first-order partial derivative of $\taubar{}$ w.r.t $\qdd$ is ${\partial \taubar}/{\partial \qdd} = \M(\q)$. Taking subsequent partial derivatives results in ${\partial ^2 \taubar}/{\partial \qdd \partial \qdd}=0$ and  ${\partial^2 \taubar}/{\partial \qdd \partial \qd}=0$. However, the cross-derivative w.r.t $\qdd$ and $\q$ is non-trivial and equals ${\partial \M(\q)}/{\partial \q}$. Garofalo et al.~\cite{garofalo2013closed} present formulas for the partial derivative of $\M(\q)$ w.r.t $\q$ for multi-DoF Lie group joints. In this work, we re-derive that result using newly developed spatial matrix operators and contribute new analytical SO partial derivatives of ID w.r.t $\q$ and $\qd$.

For single-DoF joints, each block of the Hessian tensor $\frac{\partial^2 \taubar_{i}}{\partial \u_{j} \partial \u_{k}}$ is a scalar representing a conventional SO derivative w.r.t joint angles, rates, or accelerations. In this case, the order of $\u_{j}$ and $\u_{k}$ doesn't matter. This operation becomes more nuanced when considering derivatives w.r.t configuration for a multi-DoF joint, wherein we define the operator $\frac{\partial}{\partial \q_k}$ to represent a collection of Lie derivatives, as in \cite{singh2022efficient}. For example, $\frac{\partial \taubar_{i}}{ \partial \q_k}$ is defined as the $n_i \times n_k$ matrix where each column gives the derivative of $\taubar_{i} \in \mathbb{R}^{n_i}$ w.r.t~changes in configuration along one of the $n_k$ free modes of joint $k$ (see \cite{singh2022efficient} for detail). When the Lie derivatives along the free modes of a joint do not commute, %the order of $\q_{j}$ and $\q_{k}$ matters, and 
$\frac{\partial^2 \taubar_{i} }{\partial \q_{k} \partial \q_{k}}$ can lack the usual symmetry properties. Such a case occurs, for example, with a spherical joint, since rotations do not commute. 
%\ne \frac{\partial^2 \taubar_{i}}{\partial \q_{k} \partial \q_{j}}$ in general. 
%The dynamics Hessian tensor can be visualized as a 3 dimensional array $\mathcal{Z}$ for which $\mathcal{Z}_{i,j,k}=\frac{\partial^2 \taubar_{i}}{\partial \u_{j} \partial \u_{k}}$.
 To obtain the partial derivatives of spatial quantities embedded in Eq.~\ref{dtau_dq}-\ref{dtau_dqd}, some identities (\appref{mdof_iden}) are derived. These are an extension to ones defined in Ref.~\cite{singh2022efficient}, but use the newly developed spatial matrix operators from Sec.~\ref{sva_tensor}. 
 
 For example, identity K1 (\appref{mdof_iden}) is an extension of identity J1 in Ref.~\cite{singh2022efficient}. The identity J1 (Eq.~\ref{J1_eqn}) gives the directional derivative of the joint motion sub-space matrix $\Phibar_{i}$ w.r.t the $p^{th}$ DoF of a previous joint $j$ in the connectivity tree:
\begin{equation}
\frac{\partial \Phibar_{i}}{\partial \q_{j,p}}=  \phibar_{j,p}   \times \Phibar_{i} ~~~ (j \preceq i)
\label{J1_eqn}
\end{equation}
where $\phibar_{j,p}$ is the $p$-th column of $\Phibar_j$. On the other hand, K1 uses the $\timesM$ operator to extend it to the partial derivative of $\Phibar_{i}$ w.r.t the full joint configuration $\q_{j}$ to give the tensor $\frac{\partial \Phibar_{i}}{\partial \q_{j}}$ as:
\begin{equation}
\frac{\partial \Phibar_{i}}{\partial \q_{j}}=  \Phibar_{j}   \timesM \Phibar_{i}   ~~~ (j \preceq i) 
\end{equation}
The identities K4 and K9 describe the partial derivatives of $\Psibardot_{i}$ and $\Psibarddot_{i}$ present in Eq.~\ref{dtau_dq}-\ref{dtau_dqd}. Identities K6, K10, and K12 give the partial derivatives of the composite Inertia ($\IC{i}$), body-level Coriolis matrix ($\BC{i}$), and net composite spatial force on a body ($\f_{i}^{C}$). Individual partial derivatives of these quantities allow us to use the plug-and-play approach to simplify the algebra needed for SO partial derivatives of ID.

To calculate the SO partial derivatives, we take subsequent partial derivatives of the terms $\frac{\partial \taubar_{i}}{\partial \q_{j}}$, $\frac{\partial \taubar_{j}}{\partial \q_{i}}$,$\frac{\partial \taubar_{i}}{\partial \qd_{j}}$, and  $\frac{\partial \taubar_{j}}{\partial \qd_{i}}$ w.r.t joint configuration ($\q$), and joint velocity ($\qd$) for joint $k$. 
We consider three cases, by changing the order of index $k$ as:
\[
\textbf{Case~A:}~ k \preceq j \preceq i ~~~
\textbf{Case~B:}~  j \prec k \preceq i ~~~
\textbf{Case~C:}~ j \preceq i \prec k
\]
%The sections below outline the approach taken for calculating the SO partial derivatives. Only some cases are shown to illustrate the main idea, while a detailed summary of all the cases can be found in \appref{summary_sec}, with step-by-step derivations in Ref.~\cite{arxivss_SO} for full documentation.
The sections below outline the approach for calculating the SO partial derivatives. Only some cases are shown to illustrate the main idea, with a detailed summary of all the cases in \appref{summary_sec}, and full step-by-step derivations in Ref.~\cite{arxivss_SO}. %for full documentation.

\subsection{Second-order partial derivatives w.r.t $\q$}
For SO partials of ID w.r.t $\q$, we take the partial derivatives of Eq.~\ref{dtau_dq_1}, and~\ref{dtau_dq_2} w.r.t $\q_{k}$ for cases A, B, and C mentioned above. Although, the symmetric blocks in the Hessian allow us to re-use three of those six cases.  For any of the cases, the partial derivatives of Eq.~\ref{dtau_dq_1} and~\ref{dtau_dq_2} can be taken, as long as the accompanying conditions on the equations are met. For example, for Case B ($j \prec k \preceq i$), since $j \prec i$, only Eq.~\ref{dtau_dq_2} can be used. A derivation for Case C ($j \preceq i \prec k$) is shown here as an example. We take the partial derivative of Eq.~\ref{dtau_dq_1} w.r.t $\q_{k}$. Applying the product rule, and using the identities K4, K9, and K13 as:
\begin{equation}
  \frac{\partial^2 \taubar_{i}}{\partial \q_{j} \partial \q_{k}} = 2\Phibar_{i} \T \left(  \frac{\partial \B_{i}^{C}}{\partial \q_{k}}\Psibardot{}_{j} \right) +  \Phibar_{i} \T  \left(\frac{\partial \I_{i}^{C}}{\partial \q_{k}}\Psibarddot{}_{j} \right)
\end{equation}
The quantities $\frac{\partial \B_{i}^{C}}{\partial \q_{k}}$ and $\frac{\partial \I_{i}^{C}}{\partial \q_{k}}$ are third-order tensors where the partial derivatives of matrices $\BC{i}$ and $\IC{i}$ w.r.t each DoF of joint $k$ are stacked as matrices along the pages of the tensor. Using the identities K6 and K10 then  gives:
\begin{align}
  &\frac{\partial^2 \taubar_{i}}{\partial \q_{j} \partial \q_{k}} = 2\Phibar_{i} \T \Big(   \Bicpsiidot{k}{k}+\Phibar_{k} \timesfM \B_{k}^{C}-  \B_{k}^{C}(\Phibar_{k} \timesM) \Big)\Psibardot_{j} \nonumber\\ &
  ~~~~~~~~~~~~~~~~~~~~~+ \Phibar_{i} \T  \Big( \Phibar_{k} \timesfM \I_{k}^{C}- \I_{k}^{C}(\Phibar_{k} \timesM)\Big) \Psibarddot_{j} 
    \label{SO_q_case2_eqn1}
  \end{align}
where the tensor $\Bicpsiidot{k}{k}$ is a composite calculated for the sub-tree as  $\Bicpsiidot{k}{k}= \sum_{l\succeq k} \mathcal{B}_{l} [\Psibardot_{k}] $, with:
\begin{equation}
    \mathcal{B}_{l} \big[\Psibardot_{k} \big]  = \frac{1}{2} \big[ \big(\Psibardot_{k}\timesfM \big)\I_{l} - \I_{l} \big(\Psibardot_{k} \timesM   \big) + \big(\I_{l} \Psibardot_{k} \big) \crffM    \big]
    \label{bl_psijdot_term_defn}
\end{equation}
 Eq.~\ref{bl_psijdot_term_defn} is a tensor extension of the body-level Coriolis matrix (Eq.~\ref{bk_term_defn}) with a spatial matrix argument. Expressions for other cases are in \appref{summary_sec} with full derivation in Ref.~\cite[Sec.~IV]{arxivss_SO}.
 
\subsection{Cross Second Order Partial derivatives w.r.t $\qdd$ and $\q$}
As explained before, the cross-SO partial derivatives of ID w.r.t $\qdd$ and $\q$ results in $\frac{\partial \M}{\partial \q}$. The lower-triangle of the mass matrix $\M(\q)$ for the case $j \preceq i$ is given as~\cite{Featherstone08}:
\begin{equation}
    \M_{ji} = \Phibar_{j}\T \IC{i} \Phibar_{i}
 \label{M_eqn}
\end{equation}
Since $\M(\q)$ is symmetric~\cite{Featherstone08}, $\M_{ij} = \M_{ji}\T$. We apply the three cases A, B, and C discussed before. As an example, for Case B ($j \prec k \preceq i$), we take the partial derivative of $\M_{ji}$ w.r.t $\q_{k}$ and use the product rule, along with identity K13 as:
\begin{equation}
    \frac{\partial \M_{ji}}{\partial \q_{k}} =  \Phibar_{j}\T \left( \frac{\partial \IC{i} }{\partial \q_{k}}\Phibar_{i} +\IC{i}\frac{\partial \Phibar_{i}}{\partial \q_{k}} \right)
\end{equation}

Using with identities K1 and K6, and canceling terms gives:
\begin{equation}
        \frac{\partial \M_{ji}}{\partial \q_{k}} =  \Phibar_{j}\T ( \Phibar_{k} \timesfM \I_{i}^{C})\Phibar_{i}
        \label{M_FO_eqn1}
\end{equation}
 Expressions for other cases are listed in  \appref{summary_sec}, with details of derivation at Ref.~\cite[Sec. VII]{arxivss_SO}

\subsection{Second-order partial derivatives involving $\qd$}
For SO partial derivatives w.r.t $\qd$, we take the partial derivatives of Eq.~\ref{dtau_dqd_1} and~\ref{dtau_dqd_2} w.r.t $\qd_{k}$. 
%Here, the Case A is split into two cases: 1) $k \prec j \preceq i$, and 2) $ k=j \preceq i$. This split arises due to the condition for using identity K15 (\appref{mdof_iden}), and makes the associated algebra easier to follow. A similar split also occurs for Case B into 1) $ j \prec k \prec i$, and 2) $ j \prec k = i$. The total number of cases is five and results in eight expressions, where some of the expressions exploit symmetry of the Hessian blocks. 
A list of expressions is given in  \appref{summary_sec}, with full derivation in Ref.~\cite[Sec. V]{arxivss_SO}.

%\subsection{Cross Second order partial derivatives w.r.t $\q$,$\qd$}
For cross-SO partial derivatives of ID, we take the partial derivative of Eqs.~\ref{dtau_dqd_1}-\ref{dtau_dqd_2} w.r.t $\q_{k}$ to get $\frac{\partial^2 \taubar}{\partial \qd \partial \q}$. The three cases A,B, and C (Sec.~\ref{prem_sec}) for Eq.~\ref{dtau_dqd_1}-\ref{dtau_dqd_2} result in six expressions, which are then also used for the symmetric term $\frac{\partial^2 \taubar}{\partial \q \partial \qd}$ as:
\begin{equation}
%\small
    \frac{\partial^2 \taubar}{\partial \q \partial \qd} = \left[\frac{\partial^2 \taubar}{\partial \qd \partial \q} \right]{\!\vphantom{\Big)}}\Rten
\end{equation}
Here, we solve all the three cases A, B and C for both Eq.~\ref{dtau_dqd_1} and Eq.~\ref{dtau_dqd_2}.
Pertaining to Case A ($k \preceq j \preceq i$), since $j \preceq i$, Eq.~\ref{dtau_dqd_1} can be safely used to get $ \frac{\partial^2 \taubar_{i}}{ \partial \qd_{j}\partial \q_{k}}$. However, the $j \ne i$ requirement on Eq.~\ref{dtau_dqd_2} constrains the condition in Case A to $k \preceq j \prec i$. Similarly, for Case C ($j \preceq i \prec k$), taking partial derivative of Eq.~\ref{dtau_dqd_2} results in a stricter case $j \prec i \prec k$. 
%The indices in the final expressions are switched to order $k \preceq j \preceq i$ for efficient implementation. 
Appendix \ref{summary_sec} lists the six expressions with full derivation in Ref.~\cite[Sec. VI]{arxivss_SO}.
\section{Efficient Implementation and Algorithm}
For efficient implementation of the algorithm, all the cases are converted to an index order of $k \preceq j \preceq i$. This notation is explained with the help of following two examples.

{\em Example 1:} In Eq.~\ref{SO_q_case2_eqn1}, we first we switch the indices $k$ and $j$, followed by $j$ and $i$ to get $\frac{\partial^2 \taubar_{j}}{\partial \q_{k} \partial \q_{i}}$ as
%
%\begin{equation}
\begin{align}
  &\frac{\partial^2 \taubar_{j}}{\partial \q_{k} \partial \q_{i}} = 2\Phibar_{j} \T \Big(   \Bicpsiidot{i}{i}+\Phibar_{i} \timesfM \B_{i}^{C}-   \B_{i}^{C}(\Phibar_{i} \timesM) \Big)\Psibardot_{k} \nonumber\\
  &~~~~~~~~~~~~~~~ + 
  \Phibar_{j} \T  \Big( \Phibar_{i} \timesfM \I_{i}^{C}-  \I_{i}^{C}(\Phibar_{i} \timesM)\Big) \Psibarddot_{k} \label{tau_SO_q_1C}
    %\end{aligned}
\end{align}
For the term  $\frac{\partial^2 \taubar_{j}}{\partial \q_{i} \partial \q_{k}}$, when $k \ne i$, the symmetry property of Hessian blocks can be exploited:
\begin{equation}
    \frac{\partial^2 \taubar_{j}}{\partial \q_{i} \partial \q_{k}} =  \left[  \frac{\partial^2 \taubar_{j}}{\partial \q_{k} \partial \q_{i}} \right]{\!\vphantom{\Big)}}\Rten, (k \preceq j \prec i)
      \label{tau_SO_q_2C}
\end{equation}
The 2-3 tensor rotation in Eq.~\ref{tau_SO_q_2C} occurs due to symmetry along the 2\textsuperscript{nd} and 3\textsuperscript{rd}  dimensions.

{\em Example 2:}
\noindent In Eq.~\ref{M_FO_eqn1}, switching indices $k$ and $j$ leads to the index order $k \prec j \preceq i$. Using property M5 leads to:
\begin{equation}
       \frac{\partial \M_{ki}}{\partial \q_{j}} =  \Phibar_{k}\T ((\I_{i}^{C}\Phibar_{i}) \crffM \Phibar_{j})\Rten   
     \label{M_FO_case_1B}
\end{equation}

\noindent Symmetry of $\M(\q)$ gives us $ \frac{\partial \M_{ik}}{\partial \q_{j}}$ as:
\begin{equation}
\small
      \frac{\partial \M_{ik}}{\partial \q_{j}} = \left[\frac{\partial \M_{ki}}{\partial \q_{j}}  \right]\Tten 
     \label{M_FO_case_2B}
\end{equation}

\noindent In this case, since the symmetry is along the 1\textsuperscript{st} and the 2\textsuperscript{nd} dimension of $\frac{\partial \M_{ki}}{\partial \q_{j}} $, the tensor 1-2 rotation takes place.

The expressions for SO partials of ID (\appref{summary_sec}) are first reduced to matrix and vector form to avoid tensor operations. This refactoring is due, in part, to a lack of stable tensor support in the C++ Eigen library that is often used in robotics dynamics libraries. This reduction is achieved by considering the expressions for single DoF of joints $i$, $j$, and $k$, one at a time. We explain this process with the help of two examples. % For all the matrix expressions, refer Ref.~\cite[Sec. VIII]{arxivss_SO}
%\begin{enumerate}
%    \item 

{\em Example 1}: Considering the case $k \prec j \preceq i$ the expression 
\[
\small
\frac{\partial^2 \taubar_{i}}{\partial \qd_{j} \partial \qd_{k}} =-\big[\Phibar_{j}\T (2\Bicphi{i}{i} \Phibar_{k})\Rten \big]\Tten
\]
%for $ \frac{\partial^2 \taubar_{i}}{\partial \qd_{j} \partial \qd_{k}}$ (\appref{summary_sec}) 
 is studied for the $p^{th}$, $t^{th}$, and $r^{th}$ DoFs of the joints $i$, $j$ and $k$, respectively. The tensor term $\Bicphi{i}{i}$ (defined by Eq.~\ref{bl_psijdot_term_defn}) reduces to a matrix $\Bicphip{i}$, where $\phibar_{i,p}$ is $p$-th column of $\Phibar_i$. This term represents the value $\B_i^C$ would take if all bodies in the subtree at $i$ moved with velocity $\phibar_{i,p}$.
 The above reduction enables dropping the 3D tensor rotation $(\Rten)$. The products of $\Bicphip{i}$ with column vectors $\phibar_{j,t}$ and $\phibar_{k,r}$ then provides a scalar, resulting in dropping the rotation $(\Tten)$ from above: 

\begin{equation}
\small 
    \frac{\partial^2 \taubar_{i,p}}{\partial \qd_{j,t} \partial \qd_{k,r}} =- 2\phibar_{j,t}\T\big( \Bicphip{i}\phibar_{k,r}\big) 
    \label{eq:example1}
\end{equation}
%
%\item 
{\em Example 2:}
The term $ \frac{\partial^2 \taubar_{i}}{\partial \qd_{k} \partial \qd_{j}} = \left[ \frac{\partial^2 \taubar_{i}}{\partial \qd_{j} \partial \qd_{k}} \right]\Rten$ is
evaluated for the $p^{th}$, $t^{th}$, and $r^{th}$ DoFs of the joints $i$, $j$ and $k$, respectively. In this case, the 2-3 tensor rotation ($\Rten$) drops out, since the resulting expression is a scalar.

\begin{equation}
\small
\frac{\partial^2 \taubar_{i,p}}{\partial \qd_{k,r} \partial \qd_{j,t}} =  \frac{\partial^2 \taubar_{i,p}}{\partial \qd_{j,t} \partial \qd_{k,r}}
\label{so_qd_eqn_2}
\end{equation}
%\end{enumerate}

Algorithm~\ref{alg:tau_SO_v5} (IDSVA SO) is detailed in \appref{summary_sec} and returns all the SO partials from Sec.~\ref{so_partials_sec}. 
%The algorithm has a computational complexity $\mathcal{O}(Nd^2)$ where $d$ is the depth of the kinematic tree. 
It is implemented with all kinematic and dynamic quantities represented in the ground frame. %\red{A version of the algorithm, in which the expressions for partial derivatives are reduced to matrix form, instead of a scalar form was formulated. The strategy of computing the intermediate quantities for an entire sub-tree ($\subtree(j)$) of body $j$ as used in Ref.~\cite{singh2022efficient} was used as a part of it. This approach was found computationally expensive due to the dynamic-sized vectors resulting from the size of $\subtree(j)$. The algorithm~\ref{alg:tau_SO_v5} based on scalar expressions uses all fixed size 6-vectors as intermediates, and hence was computational faster.}
%When body $i$ is a successor of $j$, instead of solving the SO expressions for joint $i$, they can be evaluated for the entire sub-tree of body $j$ at the same time. In this regard, the index order $k \preceq j \preceq i$ converts to $k \preceq j \preceq \subtree(j)$, and $k \preceq j \prec i$ to  $k \preceq j \prec \subtreeb(j)$. This ordering  results in the partial derivatives for the vectors $\taubar_{\subtree(j)}$ and $\taubar_{\subtreeb(j)}$ instead of $\taubar_{i}$, and saves an inner loop. 
%
The forward pass in Alg.~\ref{alg:tau_SO_v5} (Lines \ref{alg:forward_start}-\ref{alg:forward_end}) solves for kinematic and dynamic quantities like $\a_{i}$, $\BC{i}$, $\f{}_{i}^{C}$, $\Psibardot_{i}$, and $\Psibarddot_{i}$ for the entire tree. The quantity $\frac{1}{2}$ is skipped from the definition of $\BC{i}$ to make algebra simpler, and necessary adjustments are made in the algorithm. The backward pass then cycles from leaves to root of the tree and consists of three main nested loops, each one for bodies $i$, $j$, and $k$. These three nested loops also consist of a nested loop for each DoF of joint $i$, $j$, and $k$. The indices $p$ (from 1 to $n_i$, Line \ref{alg:1b_start}), $t$ (from 1 to $n_j$, Line \ref{alg:2b_start}), and $r$ (from 1 to $n_k$, Line \ref{alg:3b_start}) cycle over all of the DoFs of joints $i$, $j$, and $k$ respectively. 
Some of the intermediate quantities in the algorithm are defined for a DoF of a joint. For example, in Line \ref{alg:1_define}, $\phibar_{p}$  %is the Joint Subspace Column Vector for the $p^{th}$ DoF of the multi-DoF joint $i$. Hence, $\phibar_{p}$ 
is simply the $p^{th}$ column of $\Phibar_{i}$. Similar quantities for joint $j$ and $k$ are defined in Line \ref{alg:2_define} and \ref{alg:3_define} respectively. 

The backward pass (Lines \ref{alg:1b_start}-\ref{alg:1b_end}) cycles  $n$ times in total (i.e., over all joints), with the nested loops (Lines \ref{alg:2b_start}-\ref{alg:2b_end} and \ref{alg:3b_start}-\ref{alg:3b_end}) each executing at most $6d$ times per cycle of its parent loop. Thus, the total computational complexity is $O(Nd^2)$.

%The index $d$ (Alg.~\ref{alg:tau_SO_v5}, Line 13, 30) cycles over all the DoFs of the joint $j$ (i.e., from 1 to $n_j$), and similarly $c$ for joint $k$. The index $d'$ (Line 14, 32) is the index of the $d^{th}$ DoF of joint $j$ in the kinematic tree (i.e., from 1 to $n$), with $c'$ analogously for joint~$k$. 

%The algorithm also makes use of the property that $\u\T \B \v  = \textrm{Trace}(\v \u\T \B)$ for vectors $\u,\v$ and matrix $\B$. The colon $(:)$ notation represents reshaping a matrix into a vector. In this regard, if $\A = \v \u\T$, then $\u\T \B \v = \A(:)\T \B(:)$. Consider the implementation of Eq.~\ref{eq:example1} as an example. In line 27, the 6 $\times$ 6 matrix $\B_j^C[\phibar_{j,d}]$ is reshaped into a 36-vector for building columns of a 36 $\times$ $n$ matrix $\D_{3}$. This reshaping is key to the algorithm, since it allows using $\B_{i}^C[\phibar_{i,d}]$ for all $i\in \nu(j)$ at the same time to implement \eqref{eq:example1} on line 55. 
%from $\D_{3}$ in Line 43, and therefore get the derivatives for $\taubar_{\nu(j)}$.  
%The matrix transpose %(for example for Eq.~\ref{so_qd_eqn_2}) 
%is also dropped since the result $\frac{\partial^2 \taubar_{\subtree(j)}}{\partial \qd_{d'} \partial \qd_{c'}}$ is a vector. %\\[-1.5ex] 

While the algorithm is complex, an open-source {\sc Matlab} version of it can be found at ~\cite{matlabsource}, and is integrated with Featherstone's \texttt{spatial v2} library~\cite{Featherstone08}

\section{Accuracy and Performance}
A complex-step method \cite{cossette2020complex} was used to calculate the SO partials of ID  and verify the accuracy for the proposed algorithm. The complex-step approach was applied to FO derivatives of ID \cite{singh2022efficient} and verified derivatives accurate to machine precision. 

\begin{figure}[t]
\center
\includegraphics[width=8.0cm]{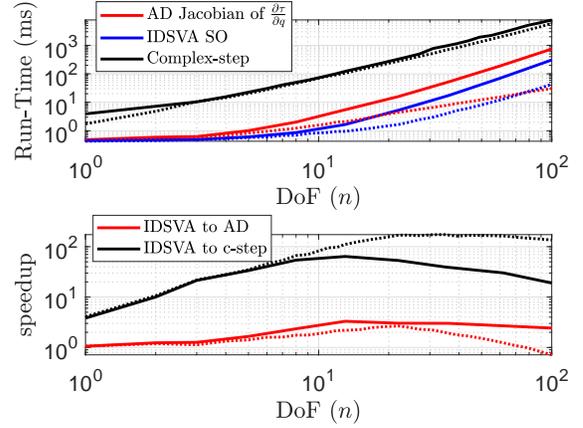}
\caption{a) IDSVA outperforms the AD $\&$ complex-step approach for all serial (bold)/branched (dashed) chains in the CasADi \cite{andersson2019casadi} virtual machine when $N<70$ b) speedup of IDSVA over AD $\&$ complex-step for serial and branched $bf=2$ chains. }
\vspace{-5px}
\label{ad_vs_idsva_cstep}
\end{figure}

For run-time comparison, the automatic differentiation tool CasADi~\cite{andersson2019casadi} in {\sc Matlab} was used. AD was used to take the Jacobian of $[\frac{\partial \taubar}{\partial \q},\frac{\partial \taubar}{\partial \qd}]\T$, via its application to the algorithm presented in Ref.~\cite{singh2022efficient} for the FO derivatives. Since CasADi is not compatible with functions defined on a Lie group, systems with single DoF revolute joints were considered. Fig.~\ref{ad_vs_idsva_cstep} shows a comparison of IDSVA with the AD and complex-step approach for serial and branched chains with a branching factor $bf$~\cite{Featherstone08}. For serial chains, IDSVA outperforms AD for all $N$, with speedups between 1.5 and 3$\times$ for models with $N>5$. For branched chains, the AD computational graph is highly efficient, resulting in performance gains beyond a critical $N$. For $bf=2$ chains, this critical $N$ lies at $N=70$. The complex-step method is accurate but slow in run-time, as seen from the plot in Fig.~\ref{ad_vs_idsva_cstep}.

\begin{figure}
\center
\includegraphics[height=5.5cm]{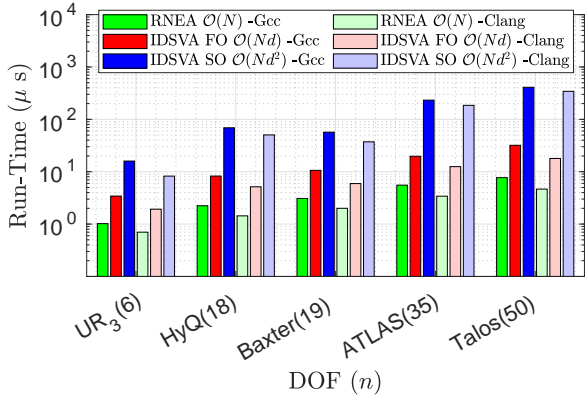}
\caption{Run-time comparison between Pinocchio~\cite{carpentier2019pinocchio} C++ implementation of RNEA~\cite{Featherstone08}, IDSVA FO~\cite{singh2022efficient}, and IDSVA SO using GCC 9.4 (dark), Clang 10.0 (light) compilers. Fixed base- $\textrm{UR}_{3}$, Baxter. Floating base- HyQ, ATLAS, Talos.}
\label{idsva}
\vspace{-5px}
\end{figure}

% Fig.~\ref{idsva_share} shows the $\%$ share of both the backward and the forward passes in IDSVA SO (Alg.~\ref{alg:tau_SO_v5}) for a branched chain with $bf=2$. For models with low $N$, both the $\mathcal{O}(N)$ forward and the $\mathcal{O}(Nd^2)$ backward passes account for almost equal run-time. But as $N$ increases, the $\%$ share of the backward pass increases such that $99\%$ of the run-time is due to the expensive $\mathcal{O}(Nd^2)$ operations. The share of run-time for each of the three SO partial derivative terms account for almost ${1}/{3}$ of the run-time of the backward pass. 
A preliminary run-time analysis was also performed with an implementation, available at \cite{cppsource}, extending the Pinocchio~\cite{carpentier2019pinocchio} open-source library.
Fig.~\ref{idsva} gives run-time numbers (in $\mu$s) for several fixed/floating base models for the IDSVA SO (\ref{alg:tau_SO_v5}) algorithm, implemented in C/C++ within the Pinocchio framework~\cite{carpentier2019pinocchio}. For reference, the run-times for RNEA~\cite{Featherstone08}, and the IDSVA FO algorithm given in \cite{singh2022efficient} are also provided. From Fig.~\ref{idsva}, the ratio of run-times for SO to FO derivatives increases with $N$ due to the algorithm complexity ratio of $d$ between the two algorithms.
All computations were performed on an Intel (R) 12th Gen i5-12400 CPU with 2.5 Ghz, with turbo boost off.
From Fig.~\ref{idsva}, SO partial derivatives of a floating base 18-DoF HyQ quadruped model take 51 $\mu$s in the C/C++ implementation.%}.

\section{Conclusions}
In this paper, SO partial derivatives of rigid-body inverse dynamics w.r.t $\q$, $\qd$, and $\qdd$ were derived using Spatial Vector Algebra (SVA) for models with multi-DoF Lie group joints. SVA was extended for spatial matrices to enable tensor operations required for the SO derivatives. This extension was done using three cross-product operators for spatial matrices. An efficient recursive algorithm was also developed to calculate the SO partial derivatives of ID by exploiting common expressions and the structure of the connectivity tree. A {\sc MATLAB} run-time comparison with AD using CasADi shows a speedup between 1.5-3$\times$ for serial chains with $N>5$. Future work will focus on closed-chain structures, more efficient C/C++ implementation, and further analysis considering code-generation and AD strategies in C/C++. 

\useRomanappendicesfalse
\appendices

\section{Multi-DoF Joint Identities}
%\vspace{-3px}
\label{mdof_iden}
The following spatial vector/matrix identities are derived by taking the partial derivative w.r.t the full joint configuration $(\q_{j})$, or joint velocity $(\qd_{j})$ of the joint $j$. In each, the
quantity to the left of the equals sign equals the
expression to the right if $j\preceq i$, and is zero otherwise,
unless otherwise stated.

%The identities apply when $j \preceq i$, and equal zero, unless otherwise stated. 
 
{
\small
\begin{align*}
&\frac{\partial \Phibar_{i}}{\partial \q_{j}}=  \Phibar_{j}   \timesM \Phibar_{i}\tag{K1}\\
&\frac{\partial \Phibardot_{i}}{\partial \q_{j}}=
    \Psibardot_{j}   \timesM \Phibar_{i}+ \Phibar_{j} \timesM \Phibardot_{i}\tag{K2}\\
&\frac{\partial (\Phibar_{i}\qd_{i} \times \Phibar_{i})}{\partial \q_{j}}=
      \Phibar_{j}   \timesM (\Phibar_{i}\qd_{i} \times \Phibar_{i})\tag{K3}\\
&\frac{\partial \Psibardot_{i}}{\partial \q_{j}} =
  \Psibardot_{j}   \timesM  \Phibar_{i} + \Phibar_{j} \timesM \Psibardot_{i}\tag{K4}\\
&\frac{\partial \I_{i}}{\partial \q_{j}}=  
    \Phibar_{j} \timesfM \I_{i}- \I_{i}(\Phibar_{j} \timesM)\tag{K5} \\
&\frac{\partial \I_{i}^{C}}{\partial \q_{j}}=
      \begin{cases}
      \Phibar_{j} \timesfM \I_{i}^{C}- \I_{i}^{C}(\Phibar_{j} \timesM), &~~~~ \text{if}\ j \preceq i \\
      \Phibar_{j} \timesfM \I_{j}^{C}- \I_{j}^{C}(\Phibar_{j} \timesM), &~~~~ \text{if}\ j \succ i 
    \end{cases} \tag{K6}\\
& \frac{\partial \a_{i}}{\partial \q_{j}}= 
    \Psibarddot_{j} - \v_{i} \times \Psibardot_{j} - \a_{i} \times \Phibar_{j}\tag{K7} \\
&\frac{\partial (\I_{i}\a_{i})}{\partial \q_{j}}= 
    \Phibar_{j}\timesfM (\I_{i} \a_{i})+ \I_{i} \Psibarddot_{j} - \I_{i}( \v_{i} \times \Psibardot_{j})\tag{K8} \\
&\frac{\partial \Psibarddot_{i}}{\partial \q_{j}}=
  \Psibarddot_{j}   \timesM  \Phibar_{i} + 2\Psibardot_{j} \timesM \Psibardot_{i}+\Phibar_{j} \timesM \Psibarddot_{i}\tag{K9} \\
&\frac{\partial \B_{i}^{C}}{\partial \q_{j}}=
     \begin{cases}
      \Bicpsiidot{i}{j}+\Phibar_{j} \timesfM \B_{i}^{C}- \B_{i}^{C}(\Phibar_{j} \timesM), & \text{if}\ j \preceq i \\
         \Bicpsiidot{j}{j}+\Phibar_{j} \timesfM \B_{j}^{C}- \B_{j}^{C}(\Phibar_{j} \timesM), & \text{if}\ j \succ i 
    \end{cases} \tag{K10}\\ 
&\frac{\partial \f_{i}}{\partial \q_{j}}=
   \I_{i} \Psibarddot_{j} + \Phibar_{j} \timesfM \f_{i} + 2 \B_{i} \Psibardot_{j} \tag{K11} \\
&\frac{\partial \f_{i}^{C}}{\partial \q_{j}}=
      \begin{cases}
      \I_{i}^{C} \Psibarddot_{j} + \Phibar_{j} \timesfM \f_{i}^{C} + 2 \B_{i}^{C} \Psibardot_{j} ,   & \text{if}\ j \preceq i \\
      \I_{j}^{C} \Psibarddot_{j} + \Phibar_{j} \timesfM \f_{j}^{C} + 2 \B_{j}^{C} \Psibardot_{j}, &\text{if}\ j \succ i 
    \end{cases} \tag{K12}\\ 
&\frac{\partial \Phibar_{i} \T}{\partial \q_{j}}=
  -\Phibar_{i} \T\Phibar_{j} \timesfM\tag{K13} \\
&\frac{\partial \Phibardot_{i}}{\partial \qd_{j}}=
  \Phibar_{j} \timesM \Phibar_{i}\tag{K14} \\
&\frac{\partial \Psibardot_{i}}{\partial \qd_{j}}=
    \begin{cases}
         \Phibar_{j} \timesM \Phibar_{i}, &{\text{if}\ j \prec i} \\
      0,& \text{otherwise}
    \end{cases}\tag{K15} \\ 
&\frac{\partial \B_{i}^{C}}{\partial \qd_{j}}=
    \begin{cases}
      \Bicphi{i}{j}, & \text{if}\ j \preceq i \\
    \Bicphi{j}{j}, & \text{if}\ j \succ i 
    \end{cases} \tag{K16} 
\end{align*}
}
%\vfill\eject

%\pagebreak

\section{Summary}
%\vspace{-5px}
\label{summary_sec}
\noindent\underline {\bf Common Terms:}
{\small %
\begin{align*}
  \red{\mathcal{A}_{1}} &\triangleq \red{\Phibar_{i} \timesfM \B_{i}^{C} - \B_{i}^{C}\Phibar_{i} \timesM}\\
    \blue{\mathcal{A}_{2}} & \triangleq \blue{\Phibar_{i} \timesfM \I_{i}^{C}   -\I_{i}^{C}   \Phibar_{i}\timesM}
\end{align*}
}

{\noindent \underline {\bf SO Partials w.r.t $\q$:}
{\small
\begin{align*}
\small	
&\frac{\partial^2 \taubar_{i}}{\partial \q_{j} \partial \q_{k}} =  -\left[\Psibardot_{j}\T [2\Bicphi{i}{i}\Psibardot_{k}]\Rten  + 2\Phibar_{j}\T ((\B_{i}^{C\T} \Phibar_{i}) \crffM\Psibardot_{k} )\Rten+\right.  \\
&~~~~~~~~~~~~~~~~~~~~~~~~~~~~~\left.\Phibar_{j}\T ((\I_{i}^{C}\Phibar_{i} ) \crffM\Psibarddot_{k} )\Rten \right]\Tten  , (k \preceq j \preceq i)\\
&\frac{\partial^2 \taubar_{i}}{\partial \q_{k} \partial \q_{j}} =  \left[\frac{\partial^2 \taubar_{i}}{\partial \q_{j} \partial \q_{k}} \right]\Rten, (k \prec j \preceq i)\\
&  \frac{\partial^2 \taubar_{k}}{\partial \q_{i} \partial \q_{j}}= \Phibar_{k}\T \Big( \big[2(\Bicpsiidot{i}{i}+\red{\mathcal{A}_{1}})\Psibardot_{j} + \blue{\mathcal{A}_{2}}\Psibarddot_{j} \big]\Rten +\\
& ~~~~~~~~~~~~~~~~\Phibar_{j} \timesfM \big( 2 \B_{i}^{C}  \Psibardot_{i} +\I_{i}^{C} \Psibarddot_{i}+\f_{i}^{C} \crff \Phibar_{i}\big) \Big)  , (k \prec j \preceq i)\\
&      \frac{\partial^2 \taubar_{k}}{\partial \q_{j} \partial \q_{i}} =  \left[\frac{\partial^2 \taubar_{k}}{\partial \q_{i} \partial \q_{j}} \right]\Rten, (k \prec j \prec i) \\
&  \frac{\partial^2 \taubar_{j}}{\partial \q_{k} \partial \q_{i}} = \Phibar_{j} \T \big( 2(  \Bicpsiidot{i}{i}+\red{\mathcal{A}_{1}}) \Psibardot_{k} +  \blue{\mathcal{A}_{2}} \Psibarddot_{k}\big) , (k \preceq j \prec i)\\
&    \frac{\partial^2 \taubar_{j}}{\partial \q_{i} \partial \q_{k}} =  \left[  \frac{\partial^2 \taubar_{j}}{\partial \q_{k} \partial \q_{i}} \right]\Rten, (k \preceq j \prec i)
\end{align*}
}

\noindent \underline {\bf SO Partials w.r.t $\qd$:}
{\small
\begin{align*}
&     \frac{\partial^2 \taubar_{i}}{\partial \qd_{j} \partial \qd_{k}} =-\big[\Phibar_{j}\T (2\Bicphi{i}{i} \Phibar_{k})\Rten \big]\Tten, (k \prec j \preceq i)\\
&     \frac{\partial^2 \taubar_{i}}{\partial \qd_{k} \partial \qd_{j}} = \left[ \frac{\partial^2 \taubar_{i}}{\partial \qd_{j} \partial \qd_{k}} \right]\Rten, (k \prec j \preceq i)\\
&     \frac{\partial^2 \taubar_{i}}{\partial \qd_{j} \partial \qd_{k}} =   -\big[ \Phibar_{j}\T  (\blue{\mathcal{A}_{2}} \Phibar_{k} )\Rten\big]\Tten , (k = j \preceq i)\\
&  \frac{\partial^2 \taubar_{k}}{\partial \qd_{i} \partial \qd_{j}} =  \Phibar_{k}\T \Big[2\Bicphi{i}{i} \Phibar_{j} \Big]\Rten, (k \prec j \prec i) \\
&     \frac{\partial^2 \taubar_{k}}{\partial \qd_{j} \partial \qd_{i}} = \left[  \frac{\partial^2 \taubar_{k}}{\partial \qd_{i} \partial \qd_{j}} \right]\Rten, (k \prec j \prec i)\\
&   \frac{\partial^2 \taubar_{k}}{\partial \qd_{i} \partial \qd_{j}} =  \Phibar_{k}\T \Big[ \big((\I_{i}^{C}\Phibar_{i})\crffM   +\Phibar_{i} \timesfM \I_{i}^{C}\big) \Phibar_{j}  \Big]\Rten , (k \prec j = i)\\
&      \frac{\partial^2 \taubar_{j}}{\partial \qd_{k} \partial \qd_{i}} = \Phibar_{j}\T \Big[2\Bicphi{i}{i} \Phibar_{k}\Big], (k \preceq j \prec i)\\
&     \frac{\partial^2 \taubar_{j}}{\partial \qd_{i} \partial \qd_{k}} = \left[   \frac{\partial^2 \taubar_{j}}{\partial \qd_{k} \partial \qd_{i}}  \right]\Rten, (k \preceq j \prec i)
\end{align*}}
}

\noindent \underline {\bf Cross SO Partials w.r.t $\q$ and $\qd$:}
{\small
\begin{align*}
\small
&  \frac{\partial^2 \taubar_{i}}{ \partial \qd_{j}\partial \q_{k}} =    -\big[ \Phibar_{j}\T \big( 2\Bicphi{i}{i} \Psibardot_{k}\big) \Rten  \big] \Tten , (k \preceq j \preceq i)\\
&    \frac{\partial^2 \taubar_{j}}{\partial \qd_{i} \partial \q_{k}} =    \Phibar_{j}\T \left[ 2\Bicphi{i}{i}\Psibardot_{k}  \right] \Rten  , (k \preceq j \prec i)\\
&  \frac{\partial^2 \taubar_{i}}{\partial \qd_{k} \partial \q_{j}} =  \big[\Phibar_{k}\T (-2\Bicphi{i}{i} \Psibardot_{j} + (2\B_{i}^{C\T} \Phibar_{i})\crffM \Phibar_{j} \\ 
&~+2(\I_{i}^{C}\Phibar_{i})\crffM \Psibardot_{j}  )\Rten 
+(\Psibardot_{k} + \Phibardot_{k} ) \T ((\I_{i}^{C}\Phibar_{i})\crffM \Psibardot_{j}  )\Rten   
      \big]\Tten  \\
&~~~~~~~~~      ,(k \prec j \preceq i)\\
&  \frac{\partial^2 \taubar_{k}}{\partial \qd_{i} \partial \q_{j}} =   \Phibar_{k}\T \left[ \big(2 \B_{i}^{C}\Phibar_{i} +
  \I_{i}^{C} ( \Psibardot_{i}  + \Phibardot_{i}) \big) \crffM \Phibar_{j}+\right.\\
&~~~~~~~~~~~~~~~~~~~~~~~~~~~~\left.2\Bicphi{i}{i}\Psibardot_{j} \right]\Rten, (k \prec j \preceq i) \\
&   \frac{\partial^2 \taubar_{j}}{\partial \qd_{k} \partial \q_{i}} = \Phibar_{j}\T \big( 2 \big(   \Bicpsiidot{i}{i}+ \red{\mathcal{A}_{1}} \big) \Phibar_{k} + \blue{\mathcal{A}_{2}}   (\Psibardot_{k} + \Phibardot_{k} ) \big),\\ &~~~~~~~~~~~~~~~~~~~~~~~~~~ (k \preceq j \prec i) \\
&  \frac{\partial^2 \taubar_{k}}{\partial \qd_{j} \partial \q_{i}} =  \Phibar_{k}\T \Big(2 \big(   \Bicpsiidot{i}{i}+\red{\mathcal{A}_{1}} \big) \Phibar_{j}+  \blue{\mathcal{A}_{2}}(\Psibardot_{j} + \Phibardot_{j} )  \Big), \\ &~~~~~~~~~~~~~~~~~~~~~~~~~~ (k \prec j \prec i)
\end{align*}
}

\noindent \underline {\bf FO Partials of $\M(\q)$ w.r.t $\q$:}
{\small
\begin{align*}
    \small
&      \frac{\partial \M_{ji}}{\partial \q_{k}}  =  \frac{\partial \M_{ij}}{\partial \q_{k}}= 0, (k \preceq j \preceq i)\\
&        \frac{\partial \M_{ki}}{\partial \q_{j}} =  \Phibar_{k}\T ((\I_{i}^{C}\Phibar_{i}) \crffM \Phibar_{j})\Rten   ,  (k \prec j \preceq i) \\
&  \frac{\partial \M_{ik}}{\partial \q_{j}} = \left[\frac{\partial \M_{ki}}{\partial \q_{j}}  \right]\Tten ,  (k \prec j \preceq i) \\
&      \frac{\partial \M_{kj}}{\partial \q_{i}} = \Phibar_{k}\T  \blue{\mathcal{A}_{2}} \Phibar_{j}, (k \preceq  j \prec i) \\
&     \frac{\partial \M_{jk}}{\partial \q_{i}} = \left[ \frac{\partial \M_{kj}}{\partial \q_{i}} \right ]\Tten  ,  (k \preceq  j \prec i)
\end{align*}
}

\bibliographystyle{IEEEtran}
%\balance
\bibliography{main.bib}

\renewcommand{\mysp}{1ex}
% Triple-Loop Algo
{\normalsize
\begin{algorithm*}[t]
\normalsize
\caption{IDSVA SO Algorithm. {\em Temporary variables in the algorithm are sized as $\A_i \in \mathbb{R}^{6 \times 6}$ for matrices, and  $\u_i \in \mathbb{R}^{6\times 1}$ for vectors.}}
\begin{multicols}{2}
\begin{algorithmic}[1]  % this [1] is used to number the lines of the 
\REQUIRE $ \q, \,\qd ,\, \qdd,\, model$
\setlength{\itemindent}{.005cm}
\addtolength{\algorithmicindent}{-.01cm}
\setstretch{1.17} %1.24
\STATE $ \v_0 = 0; \, \a_{{0}} = -\a_g  $

\FOR{$j=1$ to $N$} \label{alg:forward_start}

\STATE  $ \v_{i} =  \v_{\lambda(i)} +  \Phibar_i \qd_{i}$

\STATE $\a_{i} =  \a_{\lambda(i)} +  \Phibar_i \ddot{\q}_{i} + \v_i \times \Phibar_{i} \qd_i$ \\

\STATE $ \Phibardot_i  =  \v_{i} \times \Phibar_i  $\\[.5ex] 

\STATE $\Psibardot_i  =  \v_{\lambda(i)}\times \Phibar_i  $\\[.5ex] 
\STATE $\Psibarddot_i  =  \a_{\lambda(i)}\times \Phibar_i + \v_{\lambda(i)}\times \Psibardot_i   $ \\[.5ex]

\STATE $\I_i^C = \I_i$
\STATE $ \B_i^C =  ( \v_{i} \times^*) \I_i -  \I_i  ( \v_{i} \times ) + (\I_i \v_i )\crff   $\\[.5ex]

\STATE $ \f_i^C =  \I_i  \a_{i} + ( \v_{i} \times^*) \I_i \v_{i}  $\\[.5ex]  % 6x1 vector

\ENDFOR \label{alg:forward_end}

% Second Loop starts here

\FOR{$i=N$ to $1$} \label{alg:1a_start}

\FOR{$p=1$ to $n_{i}$} \label{alg:1b_start}
\STATE $\phibar_{p} = \Phibar_{i,p}; \psibardot_{p} = \Psibardot_{i,p}; \psibarddot_{p}=\Psibarddot_{i,p}; \phibardot_{p}=\Phibardot_{i,p}$\label{alg:1_define}

\STATE $ \B_i^C(\phibar_{p}) = ( \phibar_{p} \timesf) \IC{i} -  \IC{i}  ( \phibar_{p} \times ) + (\IC{i} \phibar_{p} )\crff   $\\
\STATE $ \B_i^C(\psibardot_{p}) =  ( \psibardot_{p} \timesf) \IC{i} -  \IC{i}  ( \psibardot_{p} \times ) + (\IC{i} \psibardot_{p} )\crff $ \\
\STATE $\A_{0} = (\IC{i}\phibar_{p})\crff  $\\
\STATE $\A_{1} = \phibar_{p}\timesf \IC{i} - \IC{i}\times  \phibar_{p} $\\
\STATE $\A_{2} = 2 \A_{0} - \B_i^C(\phibar_{p}) $\\
\STATE $\A_{3} =  \B_i^C(\psibardot_{p}) + \phibar_{p}\timesf \BC{i} - \BC{i}\times  \phibar_{p} $\\
\STATE $\A_{4} = (\BCT{i}\phibar_{p}) \crff $\\
\STATE $\A_{5} = (\BC{i}\psibardot_{p} + \IC{i}\psibarddot_{p}+\phibar_{p}\timesf\f_{i})\crff $\\
\STATE $\A_{6} = \phibar_{p}\timesf \IC{i} + \A_{0}$ \\
\STATE $\A_{7} = (\BC{i}\phibar_{p} + \IC{i}(\psibardot_{p}+\phibardot_{p}))\crff$

\STATE $j=i$
 \WHILE{$j>0$} \label{alg:2a_start}  %\\[.5ex]
 \FOR{$t=1$ to $n_{j}$} \label{alg:2b_start}
\STATE $\phibar_{t} = \Phibar_{j,t}; \psibardot_{t} = \Psibardot_{j,t}; \psibarddot_{t}=\Psibarddot_{j,t}; \phibardot_{t}=\Phibardot_{j,t}$\label{alg:2_define}

  \STATE $\u_{1} = \A_{3}\T\phibar_{t} ; \: \u_{2} =  \A_{1}\T\phibar_{t}$ 
  \STATE $\u_{3} = \A_{3}\psibardot_{t}+\A_{1} \psibarddot_{t}+\A_{5} \phibar_{t}$
  \STATE $\u_{4} = \A_{6} \phibar_{t}; \: \u_{5} = \A_{2} \psibardot_{t}+ \A_{4} \phibar_{t}$
  \STATE $\u_{6} = \B_i^C(\phibar_{p})\psibardot_{t}+\A_{7} \phibar_{t} $
  \STATE $\u_{7} = \A_{3} \phibar_{t}+\A_{1} (\psibardot_{t}+\phibardot_{t})$
  \STATE $\u_{8} = \A_{4} \phibar_{t}-\B_i^{C,T}(\phibar_{p})\psibardot{}_{t} ; \; \u_{9} = \A_{0} \phibar_{t}$
  \STATE $\u_{10} = \B_i^{C}(\phibar_{p})\phibar_{t};\: \u_{11} =\B_i^{C,T}(\phibar_{p}) \phibar_{t} $
  \STATE $ \u_{12} = \A_{1}\phibar_{t}$
  
         \STATE $k=j$

         \WHILE{$k>0$} \label{alg:3a_start} %\\[.5ex]
          \FOR{$r=1$ to $n_{k}$} \label{alg:3b_start}
            \STATE $\phibar_{r} = \Phibar_{k,r}; \psibardot_{r} = \Psibardot_{k,r}$\label{alg:3_define}
            \STATE $\psibarddot_{r}=\Psibarddot_{k,r}; \phibardot_{r}=\Phibardot_{k,r}$

             \STATE $p_{1} = \u_{11}\T \psibardot_{r}$
             \STATE $p_{2}=\u_{8}\T \psibardot_{r} +\u_{9}\T \psibarddot_{r}$
             \STATE $\dtaudqSO{i,p}{j,t}{k,r}=p_{2}; \: \dtaudqdMSO{i,p}{k,r}{j,t}=-p_{1}$ \\[\mysp]
         % j != i
              \IF{$j \neq i$}
                \STATE$\dtaudqSO{j,t}{k,r}{i,p}=\dtaudqSO{j,t}{i,p}{k,r}=\u_{1}\T \psibardot_{r}+\u_{2}\T \psibarddot_{r}$ \\[\mysp]
                \STATE$\dtaudqdMSO{j,t}{i,p}{k,r}= \u_{1}\T \phibar_{r}+\u_{2}\T (\phibardot_{r}+\psibardot_{r}) $ \\[\mysp]
                \STATE $\dtaudqdMSO{j,t}{k,r}{i,p}=p_{1}$ \\[\mysp]
              \STATE$\dtaudqdSO{j,t}{k,r}{i,p}=\dtaudqdSO{j,t}{i,p}{k,r}=\u_{11}\T \phibar_{r}$ \\[\mysp]
             \STATE$\dMdq{k,r}{j,t}{i,p}=\dMdq{j,t}{k,r}{i,p} = \phibar_{r}\T\u_{12}$  \\[\mysp]                
                
              \ENDIF
              
          % k != j
          
            \IF{$k \neq j$}
              \STATE $\dtaudqSO{i,p}{k,r}{j,t}=p_{2}; \: \dtaudqSO{k,r}{i,p}{j,t}= \phibar_{r}\T\u_{3} $ \\[\mysp]
             \STATE $\dtaudqdSO{i,p}{j,t}{k,r}=\dtaudqdSO{i,p}{k,r}{j,t}=-\u_{11}\T \phibar_{r}$\\[\mysp]
            \STATE $\dtaudqdMSO{i,p}{j,t}{k,r}= \phibar_{r}\T\u_{5}+\u_{9}\T (\phibardot_{r}+\psibardot_{r})$\\[\mysp]   
             \STATE $\dtaudqdMSO{k,r}{j,t}{i,p}=  \phibar_{r}\T\u_{6}$ \\[\mysp] \STATE$\dMdq{k,r}{i,p}{j,t}=\dMdq{i,p}{k,r}{j,t} = \phibar_{r}\T \u_{9} $\\[\mysp]

            \IF{$j \neq i$}
                \STATE $\dtaudqSO{k,r}{j,t}{i,p}=\dtaudqSO{k,r}{i,p}{j,t} $\\[\mysp] 
                \STATE $\dtaudqdMSO{k,r}{i,p}{j,t}=\phibar_{r}\T \u_{7} $\\[\mysp]
                \STATE$\dtaudqdSO{k,r}{i,p}{j,t}=\dtaudqdSO{k,r}{j,t}{i,p}=\phibar_{r}\T\u_{10} $\\[\mysp]
            \ELSE
                 \STATE $\dtaudqdSO{k,r}{j,t}{i,p}= \phibar_{r}\T \u_{4} $
            \ENDIF
            \ELSE
                \STATE $\dtaudqdSO{i,p}{j,t}{k,r}=-\u_{2}\T \phibar_{r}  $
            \ENDIF
    
        \ENDFOR \label{alg:3b_end}

          \STATE $k = \lambda(k)$ 
        
         \ENDWHILE \label{alg:3a_end}
        
     \ENDFOR \label{alg:2b_end}
  \STATE $j = \lambda(j)$ 
 \ENDWHILE \label{alg:2a_end}
    \ENDFOR \label{alg:1b_end}

    \IF {$\lambda(i) > 0$}
        \STATE $\I_{\lambda(i)}^C = \I_{\lambda(i)}^C + \I_i^C ; \, \B_{\lambda(i)}^C = \B_{\lambda(i)}^C +  \B_i^C $ \\[.5ex]
        \STATE $ \f_{\lambda(i)}^C = \f_{\lambda(i)}^C +  \f_i^C $ 
    \ENDIF
\ENDFOR \label{alg:1a_end}

\RETURN $\frac{\partial^{2}\taubar}{\partial \q^{2}},\frac{\partial^{2}\taubar}{\partial \qd^{2}},\frac{\partial^{2}\taubar}{\partial \q \partial \qd}, \frac{\partial \M}{\partial \q}$
\end{algorithmic}
\end{multicols}

\label{alg:tau_SO_v5}
\end{algorithm*}
}

\end{document}